\documentclass[10pt,twocolumn,letterpaper]{article}

\usepackage{cvpr}
\usepackage{times}
\usepackage{epsfig}
\usepackage{graphicx}
\usepackage{amsmath}
\usepackage{amssymb}
\usepackage{booktabs}
\usepackage{multirow}
\usepackage{siunitx}

\usepackage[pagebackref=true,breaklinks=true,letterpaper=true,colorlinks,bookmarks=false]{hyperref}

\cvprfinalcopy 

\def\cvprPaperID{****} 

\ifcvprfinal\fi
\begin{document}

\title{Multi-Camera Trajectory Forecasting: \\
Pedestrian Trajectory Prediction in a Network of Cameras}

\author{Olly Styles\thanks{This work was carried out while Styles was at the Rapid-Rich Object Search (ROSE) Lab, Nanyang Technological University, Singapore.},  Tanaya Guha, Victor Sanchez\\
University of Warwick\\
\tt\small {o.c.styles | tanaya.guha | v.f.sanchez-silva}\\
{\tt\small {@warwick.ac.uk}}
\and
Alex Kot\\
Nanyang Technological University\\
\tt\small{eackot}\\
{\tt\small @ntu.edu.sg}
}
\maketitle

\begin{abstract}
We introduce the task of multi-camera trajectory forecasting (MCTF), where the future trajectory of an object is predicted in a network of cameras. Prior works consider forecasting trajectories in a single camera view. Our work is the first to consider the challenging scenario of forecasting across multiple non-overlapping camera views. This has wide applicability in tasks such as re-identification and multi-target multi-camera tracking. To facilitate research in this new area, we release the \textbf{Warwick-NTU Multi-camera Forecasting Database (WNMF)}, a unique dataset of multi-camera pedestrian trajectories from a network of 15 synchronized cameras. To accurately label this large dataset (600 hours of video footage), we also develop a semi-automated annotation method. An effective MCTF model should proactively anticipate where and when a person will re-appear in the camera network. In this paper, we consider the task of predicting the next camera a pedestrian will re-appear after leaving the view of another camera, and present several baseline approaches for this. The labeled database is available online: \url{https://github.com/olly-styles/Multi-Camera-Trajectory-Forecasting}.
\end{abstract}

\section{Introduction}

\begin{figure}[t]
\begin{center}
   \includegraphics[width=\linewidth]{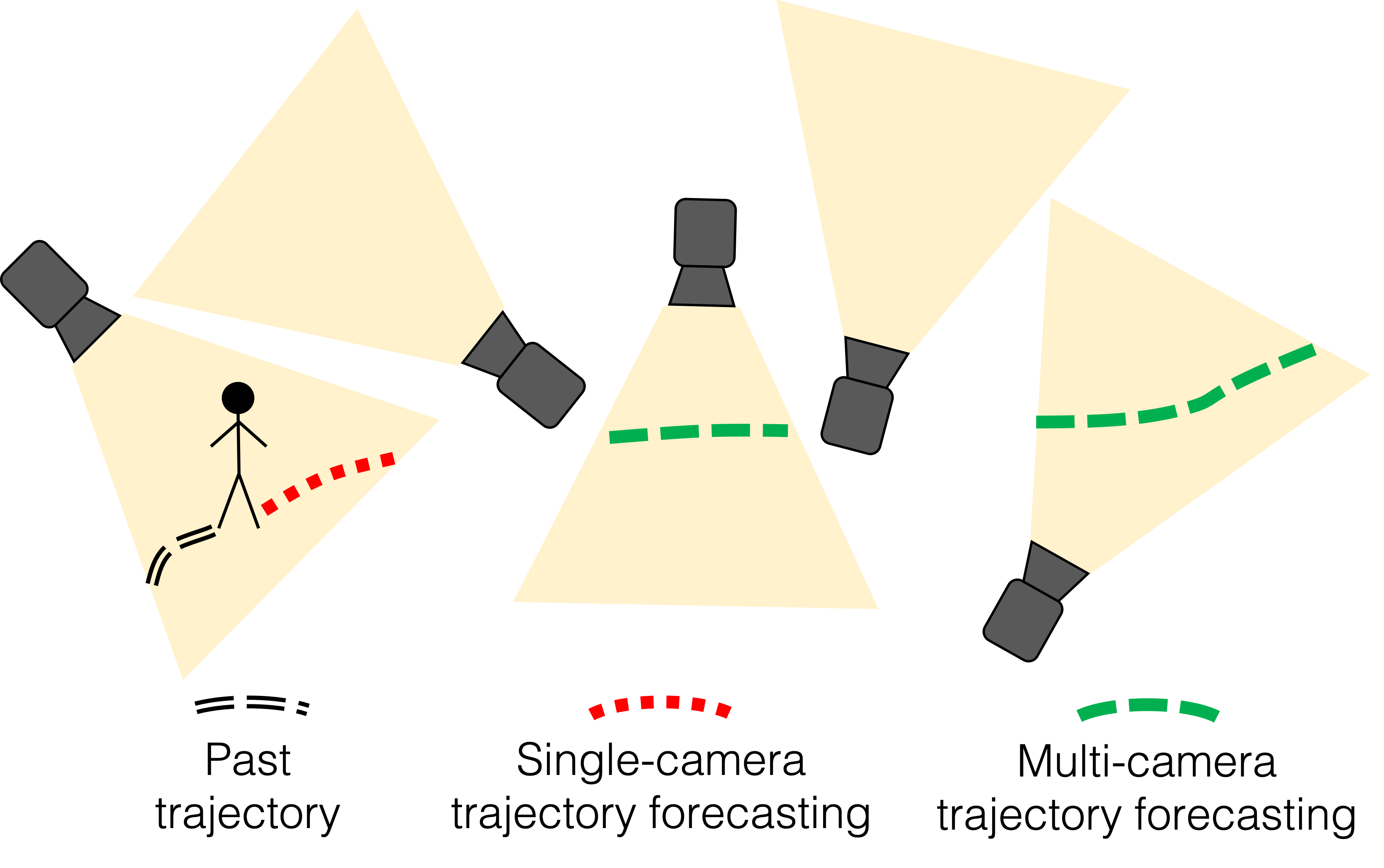}
\end{center}
   \caption{\textbf{Multi-camera trajectory forecasting (MCTF).} We introduce a novel formulation of the trajectory forecasting task which utilizes multiple camera views.}
\label{fig:pull}
\end{figure}

Predicting the future trajectory of objects in videos is a challenging problem with multiple application domains such as intelligent surveillance \cite{surveillance-survey}, person re-identification (RE-ID) \cite{spatio-temporal-reid}, and traffic monitoring \cite{multi-camera-vehicle-tracking}. Existing works on this topic focus on only the single-camera scenario, that is, predicting the future trajectory of an object in the same camera in which the object is observed \cite{sociallstm,socialgan,sophie,desire,mof}. A critical drawback of such single-camera settings is that models cannot anticipate when new objects will enter the scene. A network of multiple cameras can be used to overcome this issue. To this end, we introduce the task of \textit{multi-camera trajectory forecasting} (MCTF): Given the information about an object's location in a single camera, we want to predict its future location across the camera network, in other camera views. In particular, we want to identify the camera in which the object appears next. Fig.~\ref{fig:pull} presents an overview of the MCTF task.

\setcounter{figure}{2}
\begin{figure*}[t]
\begin{center}
   \includegraphics[width=0.9\linewidth]{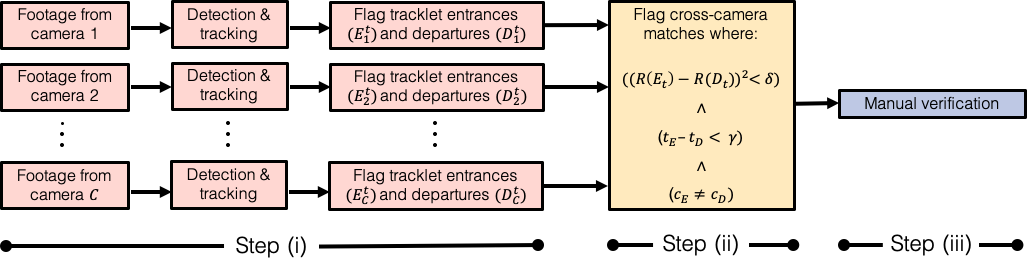}
\end{center}
\vspace{-3.5mm}
   \caption{\textbf{Annotation method.} The proposed method generates the labeled data required for MCTF with minimal human labor by using automated methods for detection, tracking, and person RE-ID before a final manual verification step.}
\label{fig:labelling}
\end{figure*}

Tracking objects (pedestrians) across a large camera network requires simultaneously running state-of-the-art algorithms for object detection, tracking, and RE-ID. Simultaneously running these algorithms can be excessively computationally demanding. Processing videos at a lower image resolution or frame-rate may reduce the computational demands, but this often results in missed detections. A successful MCTF model can address this issue by preempting the location of an object-of-interest in a distributed camera network, thereby enabling the system to monitor only selected cameras intelligently. We envision an MCTF model to be an additional component of a full multi-camera monitoring system, complementing the existing methods for detection, tracking, and RE-ID.

\setcounter{figure}{1}
\begin{figure}[t]
\begin{center}
   \includegraphics[width=\linewidth]{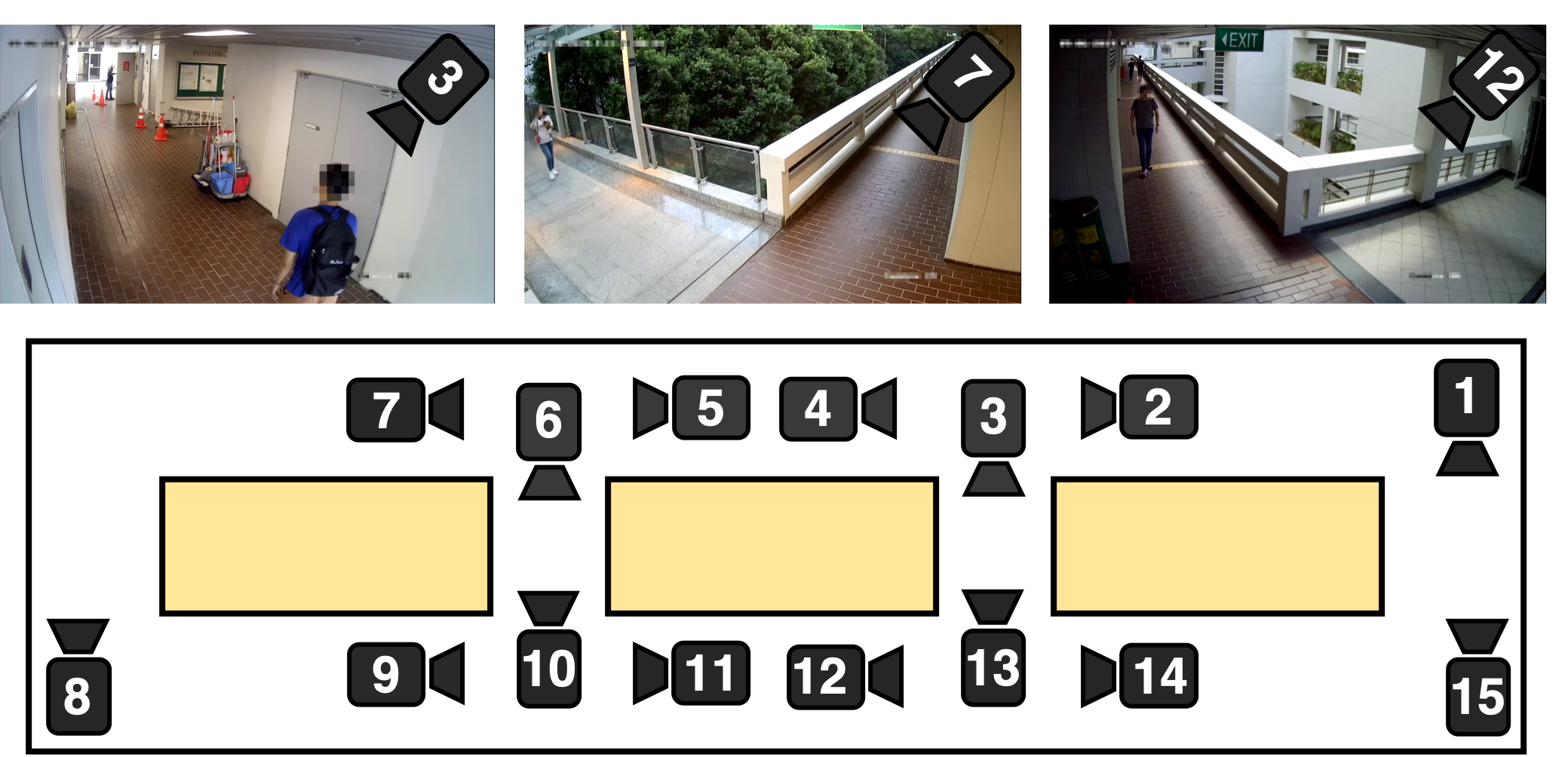}
\end{center}
\vspace{-1mm}
   \caption{\textbf{Example frames and camera network topology.} Faces have been pixelated for privacy reasons.}
\label{fig:example_frames}
\end{figure}

Trajectory information has been used previously in multi-camera settings for tasks such as person RE-ID \cite{spatio-temporal-reid} and vehicle tracking \cite{multi-camera-vehicle-tracking}. These methods, however, are \textit{reactive} to observations as they consider trajectory information to assist RE-ID or tracking only when an object has been observed in at least two cameras. In contrast, our proposed task of MCTF is \textit{proactive} - involving predicting the future location of an object even before it enters the camera view. The predicted location may then serve as a prior for the object detection algorithm, reducing the search space for detection. Owing to the wide body of complementary literature on pedestrian detection \cite{detection-survey} and RE-ID \cite{re-id-survey}, we focus on pedestrians for our MCTF task. Nevertheless, the task can be easily generalized to any moving object. To facilitate research in the newly formulated MCTF task, we collected a large dataset over 20 days using a network of 15 cameras. We present a semi-automated data annotation method that allows us to gather labels suitable for MCTF using minimal human supervision.

\section{Data collection}

Existing datasets commonly used for trajectory forecasting, such as ETH \cite{eth} and UCY \cite{ucy}, consist of just a single camera view and are therefore unsuitable for MCTF. Other datasets, such as Duke-MTMC \cite{dukedataset}, are no longer publicly available. Due to the lack of datasets suitable for MCTF, we collect a new database of 600 hours of video footage from 15 overhead mounted cameras set up indoors on the \textit{Nanyang Technological University} campus. Each camera is placed with a view of either a corridor or a junction. The footage is recorded for 20 days in 20-minute long segments collected evenly during the daytime. Example frames and the camera network topology is shown in Fig.~\ref{fig:example_frames}. We describe our semi-automated data labeling method below.

\vspace{1mm}
\noindent\textbf{Data annotation.} Fully-manual annotation of data for MCTF would be prohibitively time-consuming as trajectories must first be labeled in single-camera views and then associated across cameras. To minimize the need for manual annotation, we propose a semi-automated method that uses a combination of off-the-shelf models for detection, tracking, and person RE-ID. These results are then manually verified to ensure that proposed tracks are accurate and correct cross-camera correspondences for pedestrians are found. An overview of this annotation method is shown in Fig.~\ref{fig:labelling}, which consists of the following three steps: 

(i) We run pre-trained object detection \cite{mask-rcnn} and tracking \cite{deepsort} models to locate and track pedestrians in each of the $C$ cameras. The first 20 frames of a track
form an entrance tracklet, $E^t_c =\{ e^{t}_c, \cdots, e^{t +20}_c\}$, where $e^t_c$ is the frame at timestep $t$ in camera $c$. Similarly, the last 20 frames of a track form a departure tracklet $D^t_c=\{ d^{t - 20}_c, \cdots, d^{t}_c\}$. For departure tracklets, we assume that individuals are visible only in a single camera view. We define the camera numbers of entrance and departure tracklets as $c_E$ and $c_D$, respectively. We also define the first timestep of the entrance and departure tracklets as $t_E$ and $t_D$, respectively.  

\begin{table}[t]
\setlength{\tabcolsep}{2pt} 
\begin{center}   \caption{\textbf{WNMF dataset statistics.}} 
\vspace{-2mm}
  \begin{tabular}{c c}
    \toprule
    Hours of footage & 600  \\
    Number of cameras & 15  \\
    Collection period & 20 days \\
    Time period & 8:30am – 7:30pm \\
    Video Resolution & $1920 \times 1080$  \\
    Frames per second & 5 \\
    Cross-camera matches & 13.2K \\
    Cross-camera matches after verification & 2.3K \\
    Mean cross-camera RE-IDs per track & 2.08 \\
    \bottomrule
  \end{tabular}\label{tab:dataset-statistics}
  \end{center}
  \vspace{-2mm}
  \end{table}

\begin{figure}[t]
  \centering
    \includegraphics[width=0.8\linewidth]{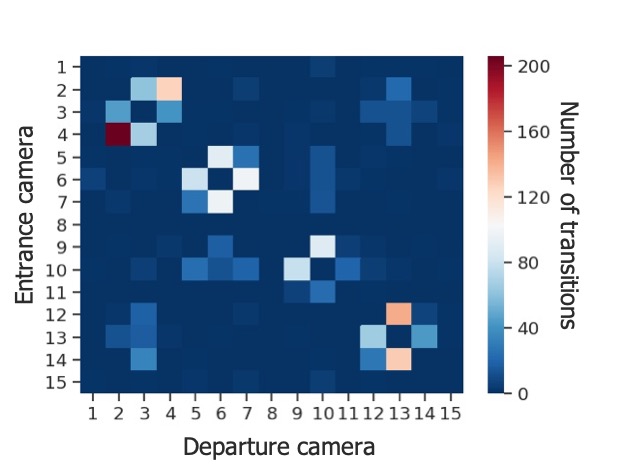}
    \caption{\textbf{Transition frequency between cameras} in the WNMF database.}
\label{fig:transition_frequency}
\end{figure}

(ii) We find cross-camera identity matches between all the departure and entrance tracklets. We use a person RE-ID model \cite{bagoftricks} to compute RE-ID features for each image and store the mean feature vector for the tracklet. We then compute the visual similarity between the entrance and departure tracklets that appear in different cameras (i.e. $c_E \neq c_D$) by computing the squared difference in their RE-ID features, $(R(E^t_c) - R(D^t_c))^2$, for all entrance and departure tracklets found in step (i), where $R(x)$ denotes the RE-ID feature vector of tracklet $x$. We retain those with a squared difference below a manually specified threshold, $\delta=0.0015$. This threshold is set deliberately high as we wish to have high recall of cross-camera matches. We are less concerned about precision, as false-positives are discarded during step (iii). In addition, we constrain candidate tracklets within a manually specified time-window $\gamma$ to cut down the search space of possible matches, i.e., we compare only those tracklets which satisfy $t_{E} - t_{D} < \gamma$. As we set $\gamma=12$ seconds, the matches are generally from neighboring cameras in the network. We confirmed this by comparing the camera transitions with respect to the network topology in Fig. \ref{fig:transition_frequency}. Our annotation method results in a set of cross-camera transitions $T = \{(E^t,D^t)\}$. 

(iii) Finally, we manually verify whether every match proposed by the algorithm is a true positive. The manual verification step assures annotation quality, as false matches and bad detections are discarded (Table \ref{tab:dataset-statistics} shows 11K such bad matches were discarded). 

As the human annotator only has to verify the cross-camera matches rather than finding them from raw videos, the manual overhead is considerably lower than fully manual data annotation. Our annotation method produces a large set (2.3K) of verified departure-entrance pairs.

\vspace{2mm}
\noindent\textbf{Data release.} The WNMF database is annotated using the aforementioned method and is available online for the research community. We provide tracking data, pre-computed RE-ID features, full multi-camera trajectories, as well as the baseline methods described in the following section. Unfortunately, we are not able to release the videos used to create this dataset due to privacy limitations.

\section{Next camera prediction} \label{sec:methods}

We evaluate the problem of predicting the next camera that a target person will re-appear, which we treat as a classification problem. The input is the past trajectory in a single camera view, and the output is a ranking that represents the next camera in the network that this person is most likely to re-appear. In this work, we focus on this next-camera prediction problem only. However, as full trajectory information is available in WNMF, it may also be used for more fine-grained MCTF in future works. 

Existing trajectory forecasting methods such as Social-LSTM \cite{sociallstm}, Social-GAN \cite{socialgan}, and SoPhie \cite{sophie} are designed for single-camera forecasting. These methods do not forecast across multiple cameras; hence direct comparison between these methods for MCTF is not possible. For a fair comparison, we instead create the following baselines:

\vspace{1mm}
\noindent\textbf{Shortest real-world distance.} We use the physical distance between cameras in the real world, and predict the camera closest to the current camera.


\vspace{0.5mm}
\noindent\textbf{Most frequent transition.} Using the transition frequency matrix computed earlier (Fig.~\ref{fig:transition_frequency}), we predict the next camera as the most frequent next camera of observation from its corresponding position. 

\vspace{0.5mm}
\noindent\textbf{Most similar trajectory.} We find the most similar trajectory in the training set to the observed trajectory, and predict the next camera to be the same as for the closest trajectory.

\vspace{0.5mm}
\noindent\textbf{Hand-crafted features.} Our hand-crafted feature vector contains velocity in $x$ and $y$ direction, acceleration in $x$ and $y$ direction, last observed bounding box height and width, and its four coordinates. We compute all features with respect to the 2D coordinate system as captured by the camera. The 10-dimensional features are classified using a single fully-connected layer.

In addition, we implement 3 purely learned approaches using the normalized bounding box coordinates as inputs. Each camera uses a separate classification network.

\vspace{0.5mm}
\noindent
\textbf{Fully connected network.} A two-layer fully connected network with 128 hidden units in each layer.

\vspace{0.5mm}
\noindent
\textbf{Long short-term memory (LSTM).} A standard LSTM with 128 hidden units.

\vspace{0.5mm}
\noindent
\textbf{Gated recurrent unit (GRU).} A standard GRU with 128 hidden units.

\section{Performance evaluation}

We compute the top 1 and top 3 classification accuracy of each method introduced in Section \ref{sec:methods}.

\vspace{0.5mm}
\noindent\textbf{Experimental setup.} We evaluate each model using 5-fold cross-validation using a challenging inter-day validation setup. Footage is recorded on different days in the validation and test sets than in the training set. In each fold, we select 10 days for training, and the remaining 5 days are split into equally sized validation and testing sets. Our neural network-based methods are each trained for 10 epochs using a batch size of 16, a learning rate of \num{1e-3}, and a dropout probability of $20\%$ between fully-connected layers.

\vspace{0.5mm}
\noindent \textbf{Discussion.} Table \ref{tab:classification} shows next camera prediction results. Predicting the correct camera in the top 3 is a straightforward problem in our dataset, given the structured camera setup and junctions with at most 3 exits. Predicting the most frequent transition using the transition matrix from the training data (Fig. \ref{fig:transition_frequency}) attains modest performance, although learned methods perform better, particularly in terms of top-1 accuracy. We suspect this is due to the past trajectory information in one camera view being informative of the person's future trajectory in a way that is not captured by other baselines. We observe moderate improvement in using recurrent models over a fully-connected network in terms of top-1 accuracy but no improvement in top-3 accuracy.

\begin{table}[tb]
\setlength{\tabcolsep}{10pt} 
\begin{center}   \caption{\textbf{Camera classification.} Given observations from one camera, the next camera of re-appearance is predicted.} 
  \begin{tabular}{l c c}
    \toprule
    \multirow{2}{*}{Model} & \multicolumn{2}{c}{Accuracy (\%)} \\
    & Top 1 & Top 3 \\
    \hline
    Shortest real-world distance & 46.8 & 92.2 \\
    Most frequent transition & 65.7 & 91.8 \\ 
    Most similar trajectory & 69.7 & 94.5 \\
    Hand-crafted features & 70.7 & 94.1 \\ 
    Fully-connected network & 73.4 & \textbf{95.1}  \\
    LSTM & 74.4 & 94.2 \\
    GRU & \textbf{75.1} & 94.9 \\
    \bottomrule
  \end{tabular}\label{tab:classification}
  \end{center}
  \vspace{-6mm}
  \end{table}
\section{Conclusion}

We have introduced a new task of human trajectory forecasting in a multi-camera scenario, which we call multi-camera trajectory forecasting (MCTF). To facilitate further research on MCTF, we presented a large dataset, WNMF, which was labeled using a semi-automated data annotation method we developed. Additionally, we presented several baseline results for predicting the next camera in which a target person re-reappears within a network. We believe our database and the preliminary results will facilitate and encourage research on this challenging problem. 

\noindent\textbf{Acknowledgements} 
This work is funded by the UK EPSRC (grant no. EP/L016400/1). The Singapore phase of this research is also partly supported by the National Research Foundation, Singapore, under its AI Singapore Programme (Award no.  AISG-100E-2018-018). Our thanks to NVIDIA for their generous hardware donation. 

{\small
\bibliographystyle{ieee_fullname}
\bibliography{egbib}
}

\end{document}